# Headline Diagnosis: Manipulation of Content Farm Headlines

Yu-Chieh Chen
Halıcıoğlu Data Science Institute
University of California San Diego
La Jolla, United States
yuc399@ucsd.edu

Pei-Yu Huang
Management and Digital Innovation
University of London
Singapore, Singapore
pyhuang001@mymail.sim.edu.sg

Chun Lin, Yi-Ting Huang, Meng Chang Chen
Institute of Information Science
Academia Sinica
Taipei, Taiwan
clin@iis.sinica.edu.tw,
ythuang@iis.sinica.edu.tw,
mcc@iis.sinica.edu.tw

## ABSTRACT

As technology grows faster, the news spreads through social media. In order to attract more readers and acquire additional profit, some news agencies reproduce massive news in a more appealing manner. Therefore, it is essential to accurately predict whether a news article is from official news agencies. This work develops a headline classification based on Convoluted Neural Network to determine credibility of a news article. The model primarily focuses on investigating key factors from headlines. These factors include word segmentation, part-of-speech tags, and sentiment features. With integrating these features into the proposed classification model, the demonstrated evaluation achieves 93.99% for accuracy.

## I. INTRODUCTION

With the rapid advancement in information technology, news agencies have expanded their platforms from newspapers to social media for greater influence. In the past, people kept current events up to date by reading newspapers and magazines. Nowadays, people gain new information from computers, phones, or tablets since publishing news on the Internet reduces money and time. Therefore, it requires less capital and technical skills to run news agencies. Some of these websites are created to profit. In order to maximize revenue and increase popularity, they spread massive amounts of advertisements along with articles on the Internet. Companies aiming for such behaviors are called "content farms."

In order to attract more readers, content farms usually reproduce massive news articles from other media in a more appealing and provocative manner. Furthermore, this news is used as propaganda for public opinions, policy, and elections. For example, new agencies manipulate original news headlines to sentimentalized ones. Since origins of these news are often unknown or untraceable, they are not as credible as those produced from the national news agencies. To help people distinguish content farm news by their headlines, a sentence-classification model based on Convoluted Neural Network is adapted. The main aspect is to find the key factors from headlines that determine whether the news is from the official news agency or content farms. As a result, word segmentations, part-of-speech tags, and sentiment scores are considered as the primary features in the model.

## II. RELATED WORK

Reference [1] focuses on analyzing incongruent headlines with their articles. Chesney et al. state that headlines on social media and press are often misleading. In order to determine clickbait, they mention that pronouns, adverbs, interrogatives, imperatives, numbers, and celebrity references are heavily used. With this information, part-of-speech is considered to be a feature in the later model. Additionally, [1] mentions that some headlines that are often classified as clickbait that do not provide information to force readers to click on the articles but use sensational technique to attract readers. Therefore, sensationalism is taken into consideration to determine whether headlines are from a content farm. Besides [1], [2] comes to the result that forward-reference in headlines are expressed by eight different manifestations of forward-reference, a stylistic technique to attract more viewers. These manifestations include demonstrative pronouns, personal pronouns, adverbs, interrogatives, definite articles, ellipsis of obligatory arguments, and imperatives and general nouns with implicit discourse references.

In order to analyze headlines, headline generation models are used to understand features of the headlines reversely. Reference [3] focuses on generating headlines based on contexts of articles with Stylistic Headline Generation (SHG) model. It generates headlines with controlled style features, including humor, romance, and clickbait. In addition to [3], to generate text, [4] focuses on content and style. The content features include theme and sentiment. The style aspects include descriptive, length, personal, and professional. However, its sentiment and professional features rely on human labelling.

As mentioned above, sensationalism is an aspect to generate sentences and analyze clickbait headlines. Therefore, a Traditional Chinese sentiment dictionary, Augmented NTU Sentiment Dictionary, is adapted to provide sentiment scores [5].

The final model that used for segregating news from content farms and national agencies are adapted from [6]. It is a sentence-classification Convoluted Neural Network, which focuses on the effects of different sizes of filters, number of feature maps, pooling strategies, and regularizations.

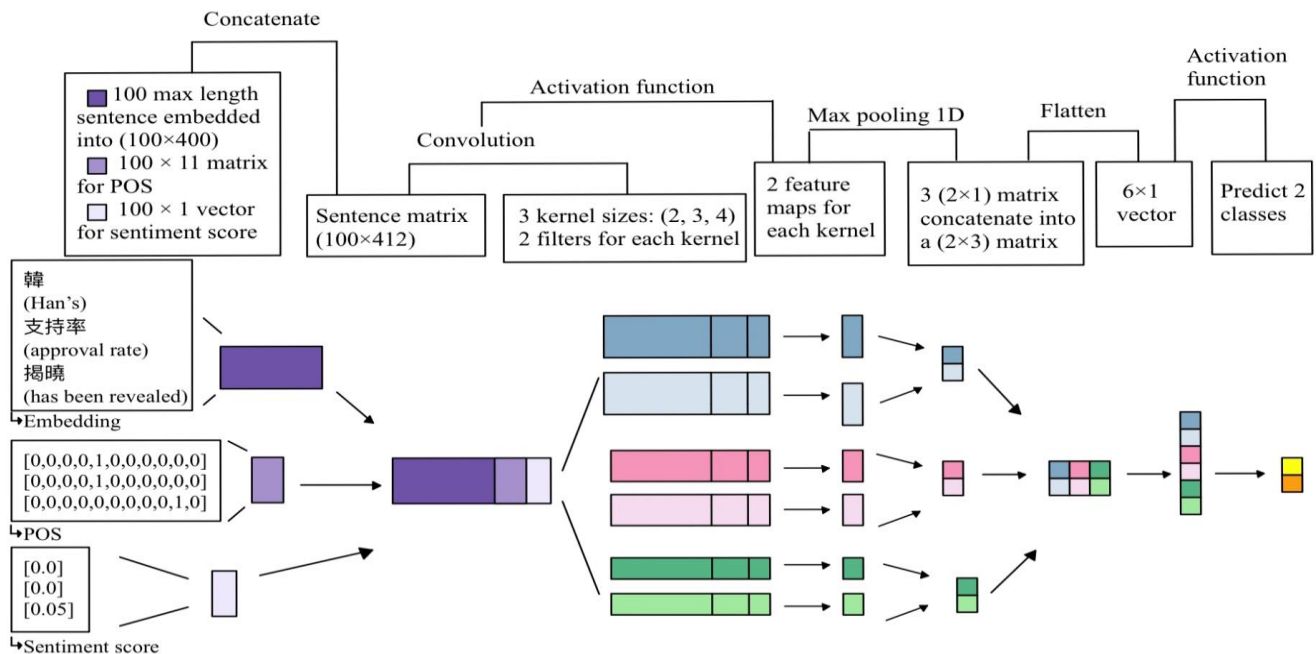

Figure 1. Overview of the Model

## III. MODEL

### A. Introduction of the Model

The basis of the model follows a sentence classification model in [6]. Reference [6] introduces a general sentence classification for NLP tasks. Generally, it is a one-layer Convoluted Neutral Network with three different kernel sizes, (2,3,4), and two filters for each kernel. 1D-max pooling is used to extract scalars from filters in the kernels. A total of three 2×1 feature vectors are concatenated into a matrix and flattened into a 6×1 matrix. In the end, an activation function is chosen to predict the binary categories. The word embedding is from [7]. It collected 655,000 words with 400 dimensions from Wikipedia in 2014. The overview of the model is shown in fig. 1.

From the baseline model to the final model, three features are proposed. In [6], Zhang and Wallace consider word embeddings as lexical feature. However, to adapt more information other than embeddings, the final model requires more features that are related to the headline, including word segmentation, part-of-speech (POS) tags, and sentiment scores. The example in fig. 1 is "韓支持率揭曉" (Han's approval rate has been revealed) is segregated into "韓" (Han's) "支持率" (approval rate) "揭曉" (has been revealed). The part-of-speech tags are provided as well. They are one-hot encoded into vectors. Besides word segmentation and part-of-speech tags, words are given with their sentiment scores. In the example above, "揭曉" (has been revealed) has a sentiment score of 0.05. The details of the features are introduced in the next section.

### B. Features

In the model, three features are used to determine the source of the news from headlines. Two attributes are from CKIP Tagger and the other is from ANTUSD. CKIP Tagger are used to provide stylistic features, including word segmentations and part-of-speech tags, and ANTUSD is for sentiment scores. Part-of-speech tags and sentiment scores are modified to fit with the models. The detailed process is shown in the following sections.

#### 1) CKIP Tagger

CKIP Tagger is an open-source word segmentation (WS), part-of-speech tagging (POS), and name-entity recognition system (NER) in Traditional Chinese. It is developed and maintained by Li and Ma [8]. It outperforms CKIPWS and Jieba-zh_TW on word segmentation. CKIP Tagger helps to separate news headlines into words with their part-of-speech tags. It offers 68 different categories of tags. For example, the part-of-speech tags and word segmentation for the headline, "華南金上半年獲利年減逾8成 每股賺0.12元" (HNFHC's profit has diminished over 80% in first two quarters; each share increases by 0.12 NTD ) are "華南金"(Nb) "上"(Nes) "半" (Neqa) "年"(Nf) "獲利"(VH) "年"(Na) "減逾"(VH) "8成"(Neqa) " "(WHITESPACE) "每"(Nes) "股"(Nf) "賺"(VC) "0.12"(Neu) "元"(Nf).

#### 2) ANTUSD:

ANTUSD (augmented NTU sentiment dictionary) collects its vocabulary from six Chinese dictionaries, including NTUSD, NTCIR MOAT task dataset, Chinese Opinion Treebank, ACBiMA, CopeOpi and E-HowNet [5]. It contains 27,221 words, and each word has a sentiment score from CopeOpi and the numbers of positive, negative, neutral, not-a-word, and non-opinion word annotations. The sentiment scores are in the range from -1 to 1. -1 is the most negative attitude and 1 is the most positive one.

## IV. EXPERIMENT

### A. Dataset

The dataset is collected from two news media, Central News Agency (CNA) and Mission (MISS). CNA is a national news agency in Taiwan. It is written in a relatively objective manner [9]. Among all other news agencies, MISS is chosen as the non-national media [10].

The reason for choosing MISS as the non-national media is that it collects news from more than 200 new agencies, including ChinaTimes, Yahoo, and ETtoday. According to [9], these three agencies are ranked as 51th, 5th, 3rd top sites in Taiwan. Besides this, MISS is known as a content farm for the following reasons. MISS modifies about 30% of headlines in the collected dataset. Among others, 50% remain the same, 12% contain broken links, and 8% do not have links. MISS changes the original headlines to a more sentimental style (table I) and uses punctuations more frequently (table II).

The CNA dataset that is collected has 22,312 news from June 3rd, 2020 to August 4th, 2020, and MISS has 43,327 news from January 4th, 2019 to August 7th, 2020. The information that are scrapped from their websites includes headline, date, passage, publisher, and original link for each news.

The average length of headline in CNA is 20.76 and 22.45 in MISS. MISS has 1.3 extra punctuations on average shown in table II. Besides punctuations, MISS has about 97% more adverbs in news headlines.

Since there are symbols such as new line, tab, and return characters in headlines, cleaning the dataset is implemented. Additionally, MISS usually collects news around different news agencies, checking the origins of the news are necessary. As a result, in 43,327 records of MISS, 61 news are found originated from CNA. Since the news from CNA are classified as credible in the research, those 61 records are eliminated.

To keep the data balanced, only 22,312 of samples are randomly selected from MISS. With a total of 44,624 data from MISS and CNA, 80% (35,699) of data are randomly chosen as training set, 10% (4,463) are used as validation set, and the other 10% (4,462) are used as testing set. Each set has equal distributed data from CNA and MISS.

TABLE I. MODIFIED MISS HEADLINE

| Version | Original Headline | MISS Headline |
|---|---|---|
| Chinese | 日列釣魚台為領土 民團批綠噤聲 | 日本列釣魚台領土 民團批民進黨連個屁都不敢放！ |
| English | Japan nationalized Pinnacle Islands (Tiaoyutai Islands) as its territory; A non-governmental organization criticizes the Democratic Progressive Party for being silent. | Japan nationalized Pinnacle Islands (Tiaoyutai Islands) as its territory; A non-governmental organization criticizes the Democratic Progressive Party would not dare to complain. |

TABLE II. POS MEAN IN A HEADLINE

| POS | CNA | MISS |
|---|---|---|
| Adjective | 0.058758 | 0.037321 |
| Conjunction | 0.080002 | 0.073557 |
| Adverb | 0.471406 | 0.931982 |
| Interjection | 0.000090 | 0.001962 |
| Noun | 5.852187 | 5.861057 |
| DE, SHI, Foreign word | 0.185909 | 0.292543 |
| Preposition | 3.335111 | 3.595956 |
| Auxiliary word | 0.002375 | 0.079142 |
| Punctuation | 0.227008 | 1.586101 |
| Verb | 0.856131 | 0.763242 |
| Whitespace | 0.160990 | 0.257391 |

TABLE III. POS FEATURES REPRESENTATIONS

| Tag | Meaning | POS Vector |
|---|---|---|
| A | Adjective | [1, 0, 0, 0, 0, 0, 0, 0, 0, 0, 0] |
| C | Conjunction | [0, 1, 0, 0, 0, 0, 0, 0, 0, 0, 0] |
| D | Adverb | [0, 0, 1, 0, 0, 0, 0, 0, 0, 0, 0] |
| I | Interjection | [0, 0, 0, 1, 0, 0, 0, 0, 0, 0, 0] |
| N | Noun | [0, 0, 0, 0, 1, 0, 0, 0, 0, 0, 0] |
| OTHER | DE, SHI, Foreign word | [0, 0, 0, 0, 0, 1, 0, 0, 0, 0, 0] |
| P | Preposition | [0, 0, 0, 0, 0, 0, 1, 0, 0, 0, 0] |
| T | Auxiliary word | [0, 0, 0, 0, 0, 0, 0, 1, 0, 0, 0] |
| PUNCT | Punctuation | [0, 0, 0, 0, 0, 0, 0, 0, 1, 0, 0] |
| V | Verb | [0, 0, 0, 0, 0, 0, 0, 0, 0, 1, 0] |
| WHITESPACE | Whitespace | [0, 0, 0, 0, 0, 0, 0, 0, 0, 0, 1] |

### B. Preprocessing

#### 1) CKIP Tagger

CKIP Tagger is used to separate the headlines into different words with their corresponding part-of-speech tags. However, it provides 68 different categories of the tags. In order to lower the dimension for part-of-speech tag, they are simplified into 11 categories. For instance, *Caa* (coordinate conjunction), *Cab* (conjunction such as "et cetera"), *Cba* (conjunction such as "in that case"), and *Cbb* (correlative conjunctions) are combined into a customized category, conjunction (C). *DE* (的 "of" or it is the word used after the adjective / 得 -ly), *SHI* (是 is), and *FW* (foreign word) were grouped as "OTHER," and all the punctuations are labeled as "PUNCT." Table III shows the customized part-of-speech tags with their abbreviations.

As illustrated in fig. 1, part-of-speech tags are transformed into 11×1 vectors. For example, "韓" (Han's) and "支持率" (approval rate) are nouns (N), so their part-of-speech vectors are both [0, 0, 0, 0, 1, 0, 0, 0, 0, 0, 0]. "揭曉" is a verb (V). Therefore, its part-of-speech vector is [0, 0, 0, 0, 0, 0, 0, 0, 0, 1, 0]. The feature vectors are shown in table III.

The preprocessing of the headlines is shown below in table IV. The headline is first separated into a list of different words then joined into a sentence, separating each word by space. Since whitespace is used as the filter of the tokenizer, to determine the difference between the actual whitespace from the original headline and the whitespace used to separate each segmented word, we changed the original whitespace to "$\alpha$." It is shown in column "Replacement of Whitespace" in table IV.

*2) ANTUSD*

The sentiment score is obtained from matching the separated words of the headlines with ANTUSD library. However, CKIP Tagger does not segment headlines perfectly into words that can be found in dictionaries. Since about 60% of vocabularies in ANTUSD have the length of two, bi-gram is used for word segmentation. Those two-character words are matched with the library again to get their corresponding sentiment scores. Then, the original segmented words are separated into individual characters with their original sentiment scores. To fill those missing scores, the words that are generated by bigram, matched with the library, and not appeared in the original segmented words are selected. The new scores are calculated back to original words with following rules:

  *a) If the original character does not have a score, then the bigram score directly represents the score of the character.*

  *b) If the original character has a score, the average of the original and new scores are calculated and replaced back to the score of the character.*

For instance, original headline is "日本單日增1584例確診創新高 東京將設武漢肺炎醫院" (Japan reaches its highest record of 1584 confirmed cases a day; Tokyo will construct COVID-19 hospital) and its word segmentation from CKIP Tagger is "日本 單日 增 1584 例 確診 創新 高 $\alpha$ 東京 將 設 武漢 肺炎 醫院". At first, the word 創新 is matched with the sentiment score of 0.382573 from ANTUSD library. By using bigram, the words and their sentiment scores, "日增" (0.0381147) and "新高" (0.0), are found. The process of generating new scores is shown in table V. In the end, it results in "日本(0) 單日(0.019057) 增(0.0381147) 1584(0) 例(0) 確診(0) 創新(0.286929) 高(0.0) $\alpha$(0) 東京(0) 將(0) 設(0) 武漢(0) 肺炎(0) 醫院(0)." As a note, in the original sentence, the word with a "X" means that it does not match with ANTUSD and does not have a sentiment score.

TABLE IV. EXAMPLE OF POS IN A HEADLINE

| Original WS | Replacement of Whitespace | Original POS | POS after replacement |
|---|---|---|---|
| ["華南金", "上", "半", "年", "獲利", "年", "減逾", "8 成", " ", "每", "股", "賺", "0.12", "元"] | 華南金 上 半 年 獲利 年 減逾 8成 $\alpha$ 每 股 賺 0.12 元 | "Nb Nes Neqa Nf VH Na VH Neqa WHITESPACE Nes Nf VC Neu Nf" | "N", "N", "N", "N", "V", "N", "V", 'N', "WHITESPACE", "N", "N", "V", "N", "N" |

TABLE V. EXAMPLE OF CALCULATION OF SENTIMENT SCORES

| Steps | Transformation of Headline |
|---|---|
| Original headline | 日本(X) 單日(X) 增(X) 1584(X) 例(X) 確診(X) 創新(0.382573) 高(X) $\alpha$(X) 東京(X) 將(X) 設(X) 武漢(X) 肺炎(X) 醫院(X) |
| 1. Separate words into characters | 日(X) 本(X) 單(X) 日(X) 增(X) 1584(X) 例(X) 確(X) 診(X) 創(0.382573) 新(0.382573) 高(X) $\alpha$(X) 東(X) 京(X) 將(X) 設(X) 武(X) 漢(X) 肺(X) 炎(X) 醫(X) 院(X) |
| 2. 日增 (0.0381147) and 新高 (0.0) are found in ANTUSD after using bigram | |
| 3. Replace "日" and "增" with its sentiment score | 日(X) 本(X) 單(X) 日(0.0381147) 增(0.0381147) 1584(X) 例(X) 確(X) 診(X) 創(0.382573) 新(0.382573) 高(X) $\alpha$(X) 東(X) 京(X) 將(X) 設(X) 武(X) 漢(X) 肺(X) 炎(X) 醫(X) 院(X) |
| 4. Replace "新" and "高" with its sentiment score | 日(X) 本(X) 單(X) 日(0.0381147) 增(0.0381147) 1584(X) 例(X) 確(X) 診(X) 創(0.382573) 新((0.382573+0)/2) 高(0.0) $\alpha$(X) 東(X) 京(X) 將(X) 設(X) 武(X) 漢(X) 肺(X) 炎(X) 醫(X) 院(X) |
| 5. Replace all the X with 0 | 日(0) 本(0) 單(0) 日(0.0381147) 增(0.0381147) 1584(0) 例(0) 確(0) 診(0) 創(0.382573) 新(0.191287) 高(0.0) $\alpha$(0) 東(0) 京(0) 將(0) 設(0) 武(0) 漢(0) 肺(0) 炎(0) 醫(0) 院(0) |
| 6. Combine scores back with original segmented words | 日本(0) 單日((0+0.0381147)/2) 增(0.0381147) 1584(0) 例(0) 確診(0) 創新((0.382573+0.191287)/2) 高(0.0) $\alpha$(0) 東京(0) 將(0) 設(0) 武漢(0) 肺炎(0) 醫院(0) |
| 7. The output headline and sentiment scores | 日本(0) 單日(0.019057) 增(0.0381147) 1584(0) 例(0) 確診(0) 創新(0.286929) 高(0.0) $\alpha$(0) 東京(0) 將(0) 設(0) 武漢(0) 肺炎(0) 醫院(0) |

## V. MODEL VARIATION

The baseline of the model has only one input, the word embedding. Then, part-of-speech tags are added to support the model. In the end, sentiment scores are taken into consideration as an attribute.

*A. Model with Word Embedding*

With the assumption that word choices are used differently in CNA and MISS, the baseline model only contains an input, the segmented words of the headline. For each headline, the max length is set as 100 words. Each word is embedded into 400 dimensions, resulting in 100×400 matrix.

*B. Model with Word Embedding and Part-of-speech Tags*

With the discovery that punctuations appear more often in MISS, the model also adapts the second input, part-of-speech tags. They are simplified from 68 categories into 11. They are also one-hot encoded to a 11×1 vector. It is concatenated with the previous embedded matrix, resulting in a size of 100×411 for each headline.

TABLE VI.  MODELS WITH SAME ACTIVATION FUNCTION

| Model | Testing Accuracy | Testing Loss |
|---|---|---|
| Word embedding | 0.9274 | 0.2533 |
| Word embedding + POS | 0.9397 | 0.2034 |
| Word embedding + POS + Sentiment | 0.9399 | 0.1968 |

TABLE VII.  MODEL WITH DIFFERENT ACTIVATION FUNCTIONS

| Activation Function | Testing Accuracy | Testing Loss |
|---|---|---|
| ReLu | 0.9352 | 0.5149 |
| Sigmoid | 0.9399 | 0.1968 |
| Tanh | 0.9388 | 0.4653 |

### C. Model with Word Embedding, Part-of-speech Tags, and Sentiment Score

To have a higher accuracy determining the difference between CNA and MISS, sentiment score is added as the third feature. By concatenating the sentiment score for each word with the embedding and part-of-speech matrices, it forms a 100×412 matrix.

## VI. RESULT

In order to improve the models, different features are added as inputs. At first, only word embedding is implemented. Later, part-of-speech tags and sentiment scores are added. Besides adding the inputs, different activation functions are experimented to maximize accuracy. Binary-cross entropy is being used as the loss function.

### A. Comparing the Models with Same Activation Function - Sigmoid

The model with word embedding, part-of-speech tags, and sentiment scores performs the best among the other two models, with the highest testing accuracy of 0.9399 and the lowest loss of 0.1968 shown in table VI. The model improves the most when part-of-speech is added.

### B. Comparing the Activation Functions with the Final Model

Among different activation functions, sigmoid outperforms others. While tanh and sigmoid do not differ too much in testing accuracy, tanh's loss is more than twice of sigmoid's loss shown in table VII.

## VII. CONCLUSION

The purpose of this research is to determine whether a news article is from the national news agency, CNA, or the content farm, MISS, by its headline. It is to provide readers information whether the news is credible. To automatically classify the label of the news, a Convoluted Neural Network headline classification is proposed. It is a one-layer model that has two filters and three kernel sizes. The inputs of the models are the embeddings of words in 400 dimensions, part-of-speech tags in 11 dimensions, and a one-dimension sentiment score. Words segmentations are used from CKIP Tagger. CKIP Tagger also provides 68 different categories of part-of-speech tags. In order to simplify the dimensions, the tags are combined into 11 groups, adjective, conjunction, adverb, interjection, noun, preposition, verb, auxiliary word, punctuations, whitespace, and others. Besides the word segmentation and part-of-speech tags, sentiment score is used as a feature in the model. The sentiment scores are found in ANTUSD library. However, most of the words do not matched with the library. Therefore, bigram is used to segregate headlines into two-character words and find their scores again in the library. Later, an average of the scores are adapted. In the end, with sigmoid activation function and three inputs, the model results in a highest accuracy score of 93.99%.

Sentiments and Styles of the news, including formats and semantics, change over time. Those changes are often the results of alterations in the environments, such as political, medical, technical, and financial fields. The biggest difference is that people change the way they read the news. In the past, people usually relied on newspapers to understand things happening in the world. However, nowadays, people receive new information by their computers or phones. Therefore, headlines of the news may change upon that. For example, because of the convenience and abundance of spreading information, news headlines are curtailed to make the readers easy to adapt news. With the uncertainties in future, it is hard to exploit the same model and features to determine the credibility of the news articles from their headlines. In order to continue the investigations, analysis of the contemporary news headlines is required.